\documentclass[sigconf,nonacm]{acmart}

\renewcommand\footnotetextcopyrightpermission[1]{}
\usepackage{booktabs}
\usepackage{tikz}
\usepackage{pgfplots}
\usepackage{graphicx}
\usepackage{subcaption}
\usepackage{dblfloatfix}
\usepackage{placeins}
\pgfplotsset{compat=1.17}

\graphicspath{{./paper_images/}} 

\begin{document}
\citestyle{acmnumeric}
\raggedbottom

\title{Sim2Radar: Toward Bridging the Radar Sim-to-Real Gap with VLM-Guided Scene Reconstruction}

\author{Emily Bejerano}
\affiliation{%
  \institution{Columbia University}
  \city{New York, NY}
  \country{USA}}
\email{eg3205@columbia.edu}

\author{Federico Tondolo}
\affiliation{%
  \institution{Columbia University}
  \city{New York, NY}
  \country{USA}}
\email{ft2505@columbia.edu}

\author{Ayaan Qayyum}
\affiliation{%
  \institution{Columbia University}
  \city{New York, NY}
  \country{USA}}
\email{aaq2109@columbia.edu}

\author{Xiaofan Yu}
\affiliation{%
  \institution{University of California, Merced}
  \city{Merced, CA}
  \country{USA}}
\email{xiaofanyu@ucmerced.edu}

\author{Xiaofan Jiang}
\affiliation{%
  \institution{Columbia University}
  \city{New York, NY}
  \country{USA}}
\email{jiang@ee.columbia.edu}

\renewcommand{\shortauthors}{Bejerano et al.}

\begin{teaserfigure}
\centering
\includegraphics[width=\textwidth]{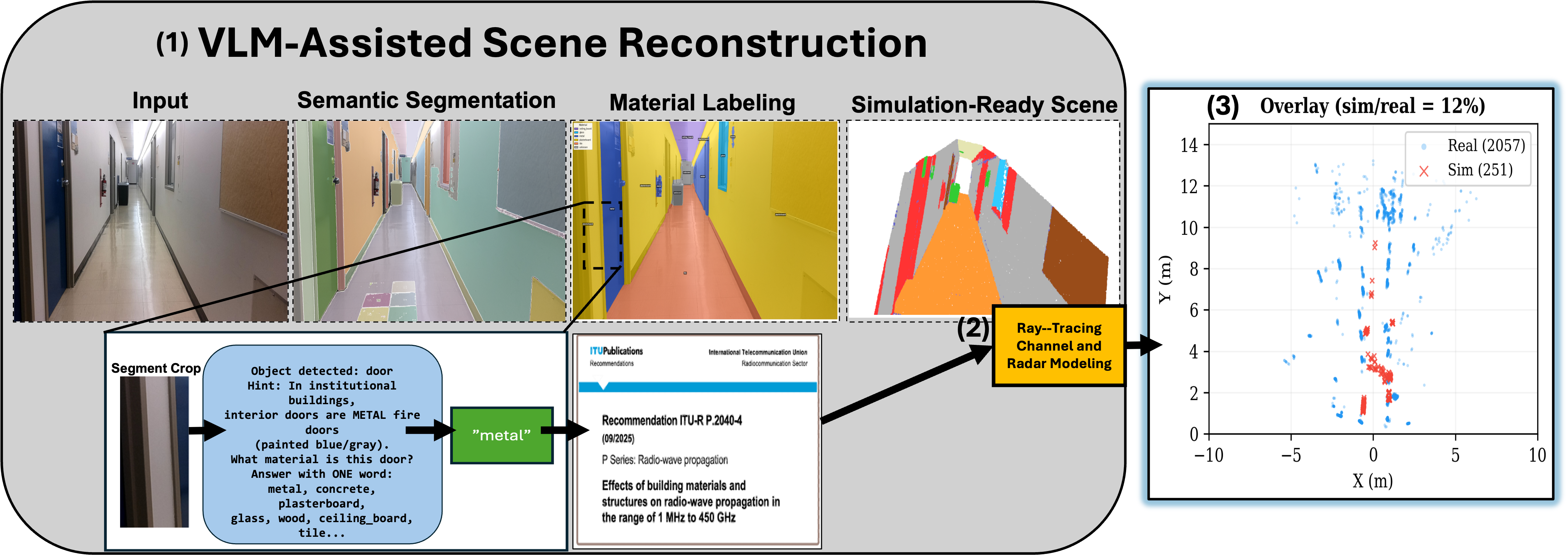}
\caption{\textbf{Sim2Radar pipeline.} (1) \textit{VLM Scene Construction}: Given an indoor RGB image, we reconstruct a material-labeled 3D mesh using monocular depth estimation, visual segmentation, and vision-language reasoning to infer material properties. (2) \textit{RT Channel Modeling}: The material-aware scene is processed by a configurable Mitsuba-based ray tracer that simulates mmWave propagation and reflection using Fresnel coefficients from ITU-R P.2040. (3) \textit{Signal Processing}: Ray tracing outputs are binned by range, azimuth, and elevation to produce synthetic radar point clouds. The resulting simulated points (red) are sparser than real measurements (blue), yet pre-training on this data improves 3D localization when fine-tuning on limited real data.}
\label{fig:pipeline}
\end{teaserfigure}

\begin{abstract}
Millimeter-wave (mmWave) radar provides reliable perception in visually degraded indoor environments (e.g., smoke, dust, and low light), but learning-based radar perception is bottlenecked by the scarcity and cost of collecting and annotating large-scale radar datasets. We present \textbf{Sim2Radar}, an end-to-end framework that synthesizes training radar data directly from single-view RGB images, enabling scalable data generation without manual scene modeling. Sim2Radar reconstructs a material-aware 3D scene by combining monocular depth estimation, segmentation, and vision-language reasoning to infer object materials, then simulates mmWave propagation with a configurable physics-based ray tracer using Fresnel reflection models parameterized by ITU-R electromagnetic properties. Evaluated on real-world indoor scenes, Sim2Radar improves downstream 3D radar perception via transfer learning: pre-training a radar point-cloud object detection model on synthetic data and fine-tuning on real radar yields up to \textbf{+3.7} 3D AP (IoU 0.3), with gains driven primarily by improved spatial localization. These results suggest that physics-based, vision-driven radar simulation can provide effective geometric priors for radar learning and measurably improve performance under limited real-data supervision.
\end{abstract}

\maketitle

\section{INTRODUCTION}

Millimeter-wave (mmWave) radar enables perception in conditions where cameras and LiDAR fail. Radar operates effectively through smoke-filled corridors, low-light indoor spaces, fog, dust, and cluttered environments where optical sensors cannot provide reliable measurements. This robustness makes radar critical for indoor robotics applications, including search-and-rescue systems, service robots navigating dynamic environments, and mobile platforms operating under degraded visual conditions, as well as autonomous vehicles in adverse weather.

Despite this promise, learning-based radar perception remains far behind visual perception. The fundamental bottleneck is \textbf{data scarcity}. Visual datasets contain millions of labeled images. Radar datasets are orders of magnitude smaller due to specialized hardware, multi-sensor synchronization complexity, and tedious annotation of sparse returns. As a result, radar perception models often overfit to narrow environments and fail to generalize.

\textbf{Simulation as a solution.} Physics-based simulation offers a path forward by modeling how electromagnetic waves interact with scene geometry and materials. If material properties can be inferred from visual inputs, synthetic radar data could be generated at scale, bypassing the need for expensive hardware and manual annotation.

\textbf{The sim-to-real challenge.} Bridging the gap between simulated and real radar remains challenging due to systematic differences in point density, noise statistics, and spatial coverage. As a result, simulated radar rarely matches real sensor measurements at the signal or point-cloud level. Naively combining simulated and real data during training can therefore degrade performance, as models may overfit to simulation-specific artifacts rather than learning task-relevant physical structure.

Our contributions can be summarized as follows:
\begin{itemize}
\item We introduce Sim2Radar, a scalable end-to-end framework that synthesizes mmWave radar training data directly from RGB images. Unlike existing approaches that require computer-aided design (CAD) models or manual scene annotation, our pipeline leverages vision-language model (VLM) reasoning to infer material electromagnetic properties from visual context alone.
\item We show that physics-based simulation, despite producing visually different measurements than real sensors, can improve 3D localization when used for pre-training. Pre-training on simulated frames improves 3D detection Average Precision (AP) by up to 3.7 points when fine-tuning on limited real data, suggesting that simulation provides useful priors even under substantial domain shift.
\item We evaluate on the Indoor FireRescue Radar (IFR) dataset and provide analysis of the sim-to-real gap, showing that improvements concentrate in spatial localization rather than detection confidence. The modular pipeline enables downstream applications including data augmentation, multi-task learning, contrastive learning for sim-to-real domain alignment, and radar-text alignment for cross-modal understanding.
\end{itemize}

\section{RELATED WORK}

\subsection{Radar Simulation}

Physics-based radar simulation traditionally requires CAD models with known material properties, limiting scalability. MATLAB's Radar Toolbox and Phased Array System Toolbox provide comprehensive frequency-modulated continuous-wave (FMCW) radar modeling capabilities, but require manual scene construction with explicit geometry and material definitions, making large-scale data generation impractical. RadSimReal~\cite{bialer2024radsimreal} bridges the gap between synthetic and real data in radar object detection by training automotive radar models on synthetic data, but assumes detailed geometry and known materials. Sionna RT~\cite{hoydis2023sionna} provides differentiable ray tracing for radio propagation modeling, enabling gradient-based optimization of wireless system parameters.

C-Shenron~\cite{srivastava2025cshenron} integrates a realistic radar simulation framework into the CARLA driving simulator by fusing LiDAR and camera data via mmWave surface scattering models, enabling large-scale synthetic radar dataset generation for autonomous driving research. DT-RaDaR~\cite{amatare2024dtradar} leverages digital twin environments with radio frequency (RF) ray-tracing for privacy-preserving robot navigation, demonstrating how simulated RF data can enable autonomous system development without camera or LiDAR sensors.

Our work differs by using VLM-assisted reconstruction to infer materials directly from RGB images, requiring no existing radar data or detailed CAD models.

\subsection{Generative AI For RF Sensing}

Recent surveys~\cite{wang2025genairf} identify key challenges in traditional RF sensing, including noise, interference, incomplete data, and high deployment costs, and demonstrate how generative models can synthesize high-quality data, enhance signal quality, and integrate multi-modal information for tasks including environment reconstruction, localization, and imaging.

RF-Genesis~\cite{chen2023rfgenesis} presents a cross-modal framework for synthesizing mmWave sensing data that combines physics-based ray tracing with diffusion models, using a custom ray tracing simulator for RF propagation while leveraging diffusion models to generate diverse 3D scenes. RF-Diffusion~\cite{chi2024rfdiffusion} adapts diffusion models specifically to the RF domain by introducing Time-Frequency Diffusion theory that captures information across time, frequency, and complex-valued domains of RF signals.

Our work focuses on material-aware scene reconstruction from RGB images, enabling radar simulation without requiring existing RF data or manual scene modeling.

\subsection{mmWave Perception And Object Recognition}

Prior work has demonstrated accurate 3D object recognition on commodity mmWave devices~\cite{he2023fusang} by leveraging the large bandwidth of mmWave radars to capture fine-grained reflected responses from object shapes. Cross-modal approaches such as mmCLIP~\cite{cao2024mmclip} align mmWave signals with text for zero-shot human activity recognition (HAR), demonstrating the potential of cross-modal learning for radar perception.

\subsection{Why VLMs For Material Classification?}

Traditional computer vision classifies materials by texture. However, material composition often requires semantic reasoning. Industrial doors in fire corridors are metal due to fire code requirements, not wood. School floor tiles are typically ceramic, not vinyl. Fire extinguishers are metal regardless of paint color. Gray surfaces could be aluminum, painted wood, or plastic, and texture alone cannot distinguish these. VLMs combine visual recognition with world knowledge about object composition, enabling material inference from semantic context rather than appearance alone.

\subsection{Indoor Radar Dataset}

Real-world radar datasets for indoor perception remain scarce. The IFR dataset~\cite{duan2025ifr} provides 27K frames of synchronized RGB camera, LiDAR, and 4D mmWave radar captured across 10 buildings with 3D bounding box annotations for doors and obstacles. The dataset uses a TI 76-81 GHz MMWCAS-RF cascade imaging radar with 12 transmit and 16 receive antenna elements. Unlike outdoor-focused datasets, IFR targets fire rescue scenarios where radar's ability to see through smoke is critical. We use IFR to evaluate our sim-to-real transfer approach.

\section{VLM-ASSISTED SCENE RECONSTRUCTION}

\begin{figure*}[t]
\centering
\begin{subfigure}[t]{0.32\textwidth}
\centering
\includegraphics[width=\textwidth]{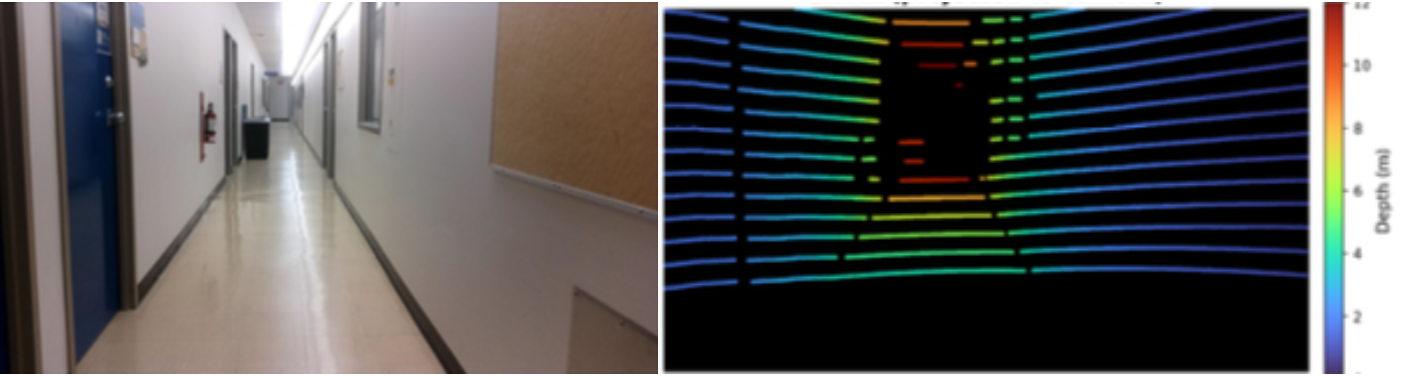}
\caption{Input data}
\label{fig:input}
\end{subfigure}
\hfill
\begin{subfigure}[t]{0.32\textwidth}
\centering
\includegraphics[width=\textwidth]{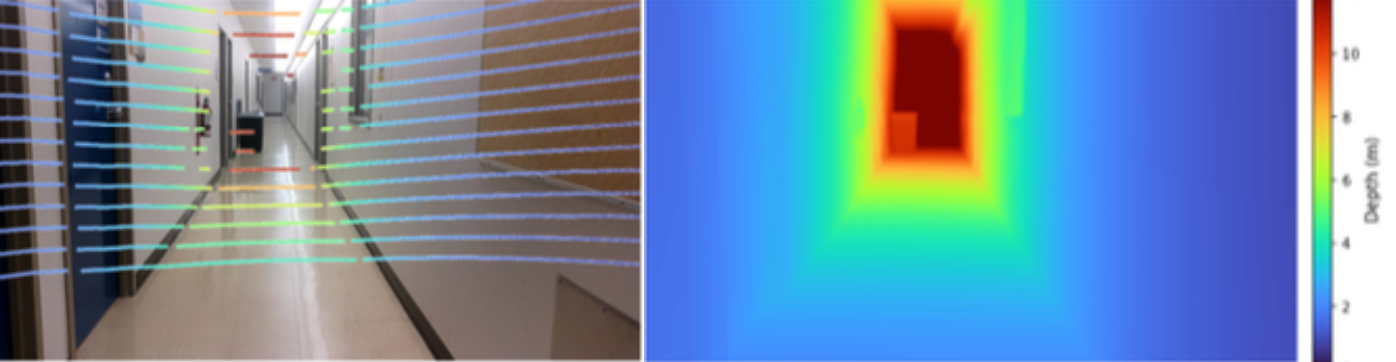}
\caption{Depth estimation}
\label{fig:depth}
\end{subfigure}
\hfill
\begin{subfigure}[t]{0.32\textwidth}
\centering
\includegraphics[width=\textwidth]{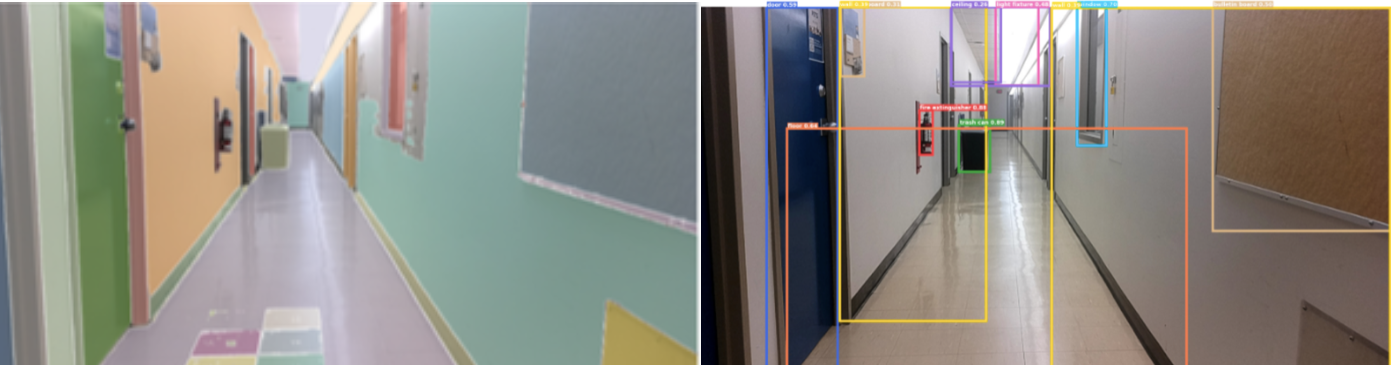}
\caption{Segmentation}
\label{fig:segmentation}
\end{subfigure}

\vspace{0.5em}

\begin{subfigure}[t]{0.32\textwidth}
\centering
\includegraphics[width=\textwidth]{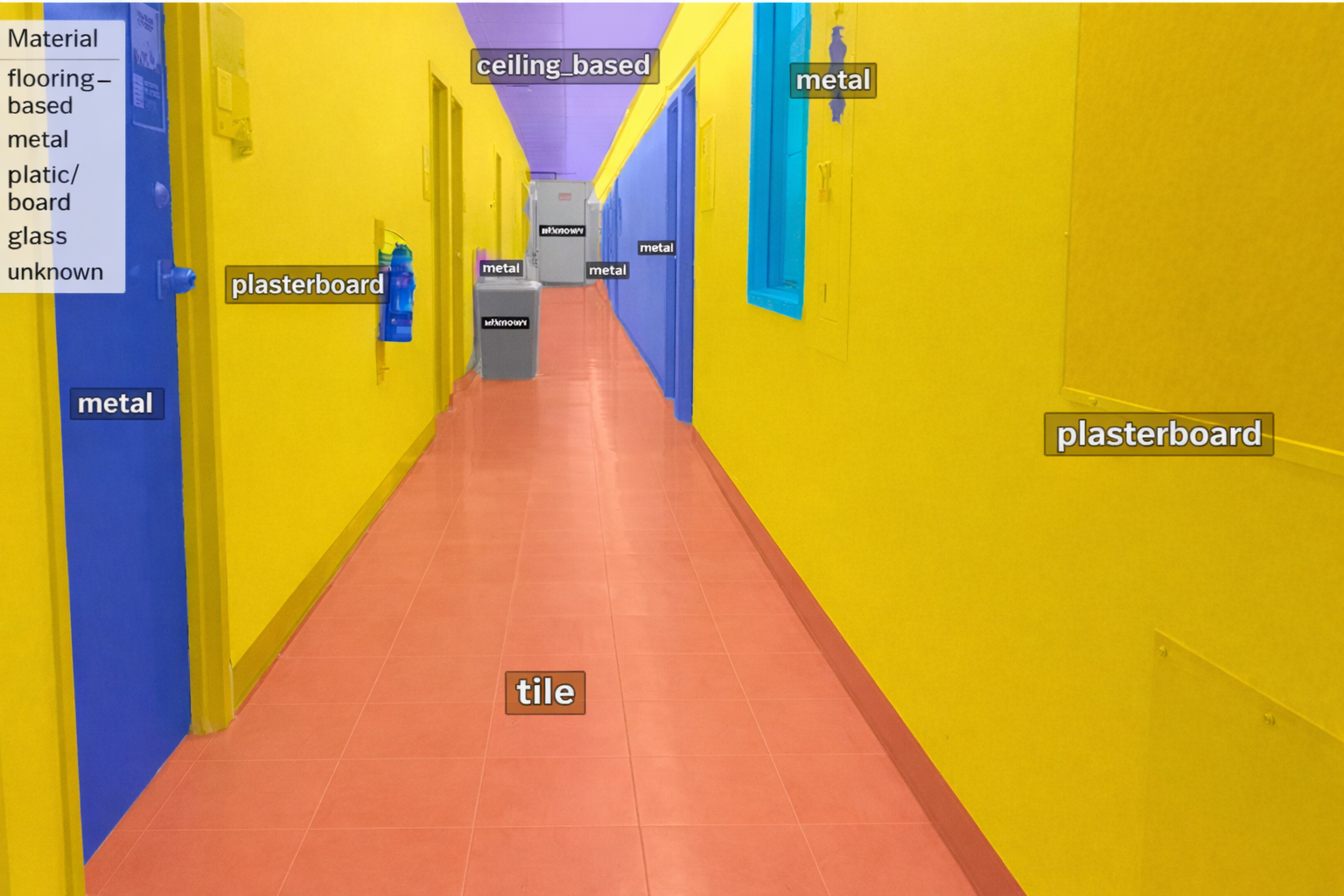}
\caption{Material classification}
\label{fig:materials}
\end{subfigure}
\hspace{2em}
\begin{subfigure}[t]{0.32\textwidth}
\centering
\includegraphics[width=\textwidth]{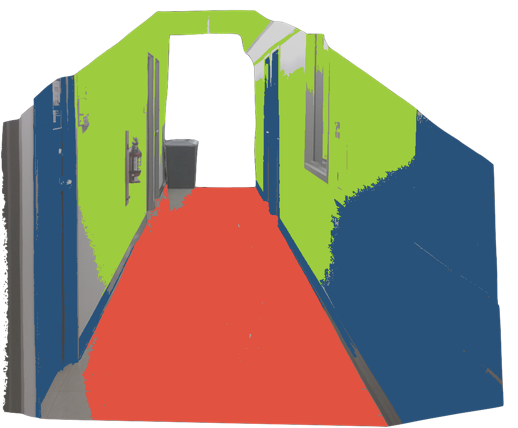}
\caption{3D scene output}
\label{fig:3dscene}
\end{subfigure}
\caption{\textbf{VLM-assisted scene reconstruction pipeline.} (a) RGB and sparse LiDAR input. (b) Dense depth from MoGe. (c) SAM2 masks and Grounding DINO detections. (d) InternVL2.5-8B material labels. (e) Material-labeled 3D scene ready for ray tracing.}
\label{fig:pipeline_steps}
\end{figure*}

Our reconstruction pipeline transforms RGB images into material-labeled 3D scenes through four stages: depth estimation, segmentation, material classification, and 3D projection. Figure~\ref{fig:pipeline} shows the complete system and Figure~\ref{fig:pipeline_steps} illustrates each stage.

\subsection{Input Requirements}

The pipeline accepts RGB images as the primary input. When available, sparse LiDAR provides metric scale for the reconstructed geometry.

\subsection{Stage 1: Monocular Depth Estimation}

MoGe v2~\cite{wang2025moge} predicts dense depth maps from RGB images. MoGe captures fine geometric details including thin structures (handrails, door frames) and small objects (fire extinguishers) that sparse LiDAR misses.

When LiDAR is available, we perform scale-and-shift alignment to ground the monocular depth in metric coordinates:
\begin{equation}
s^*, t^* = \arg\min_{s,t} \sum_i (s \cdot d_{\text{mono}}(p_i) + t - d_{\text{lidar}}(p_i))^2
\end{equation}

\subsection{Stage 2: Automatic Segmentation And Detection}

SAM2~\cite{ravi2024sam2} segments the scene into coherent regions. Grounding DINO~\cite{liu2023groundingdino} detects and labels objects using a text prompt covering indoor elements: ``door, wall, floor, ceiling, window, fire extinguisher, handrail.'' Each SAM2 mask is matched to Grounding DINO detections via Intersection over Union (IoU).

\subsection{Stage 3: VLM Material Classification}

InternVL2.5-8B~\cite{chen2024internvl} classifies each segment into material categories: metal, glass, wood, plasterboard, ceramic tile, concrete, fabric, plastic, or unknown. We select InternVL2.5-8B based on its strong performance on material property prediction in PhysBench~\cite{chow2025physbench}, a benchmark for evaluating physical world understanding in VLMs. The model is also computationally efficient, enabling single-GPU deployment.

VLMs outperform texture-based methods because they reason about object-material relationships. An industrial door is inferred to be metal based on fire code knowledge, not visual texture.

\subsection{Stage 4: 3D Projection And Output}

The final stage projects 2D segments with depth into 3D point clouds. Each pixel with a valid depth estimate is back-projected into 3D space using camera intrinsics, producing a material-labeled point cloud. The segmented surfaces are then converted into meshes and associated with material models that encode how millimeter-wave signals interact with each surface. The output is a complete, ray-tracer-ready 3D scene with per-point material labels and metric geometry.

\section{PHYSICS-BASED RADAR SIMULATION}

Given a material-labeled 3D scene, we simulate radar measurements using ray tracing with physically accurate reflection models. The simulation pipeline is modular and configurable: users can specify radar parameters from hardware datasheets including carrier frequency, bandwidth, chirp configuration, and antenna array geometry. The scene reconstruction and ray tracing components are decoupled, allowing the same reconstructed environment to be used with different radar configurations. For this evaluation, we configure the simulator to match the IFR dataset's TI MMWCAS-RF cascade radar parameters (76-81 GHz, 0.038m range resolution, 1.18$^\circ$ azimuth resolution).

\subsection{Material Electromagnetic Properties}

Each material category maps to electromagnetic properties at 77 GHz using empirically measured values from ITU-R P.2040-4. The mapping assigns permittivity and conductivity values to materials including metal, glass, ceramic tile, concrete, plasterboard, and wood, which determine Fresnel reflection coefficients at each surface. These coefficients govern how much transmitted energy is reflected back to the radar, with metals producing strong returns and low-permittivity materials like wood producing weak returns.

\subsection{Ray Tracing With Mitsuba}

We use Mitsuba 3~\cite{jakob2022mitsuba3} for simulation. The process casts rays from the radar position across the field of view, computes ray-mesh intersections, and calculates Fresnel reflection amplitudes from material properties. We accumulate multi-bounce reflections up to 2 bounces and bin returns by range, azimuth, and elevation with coherent summation.

\subsection{Analysis Of The Sim-Real Gap}

\begin{figure}[t]
\centering
\includegraphics[width=\columnwidth]{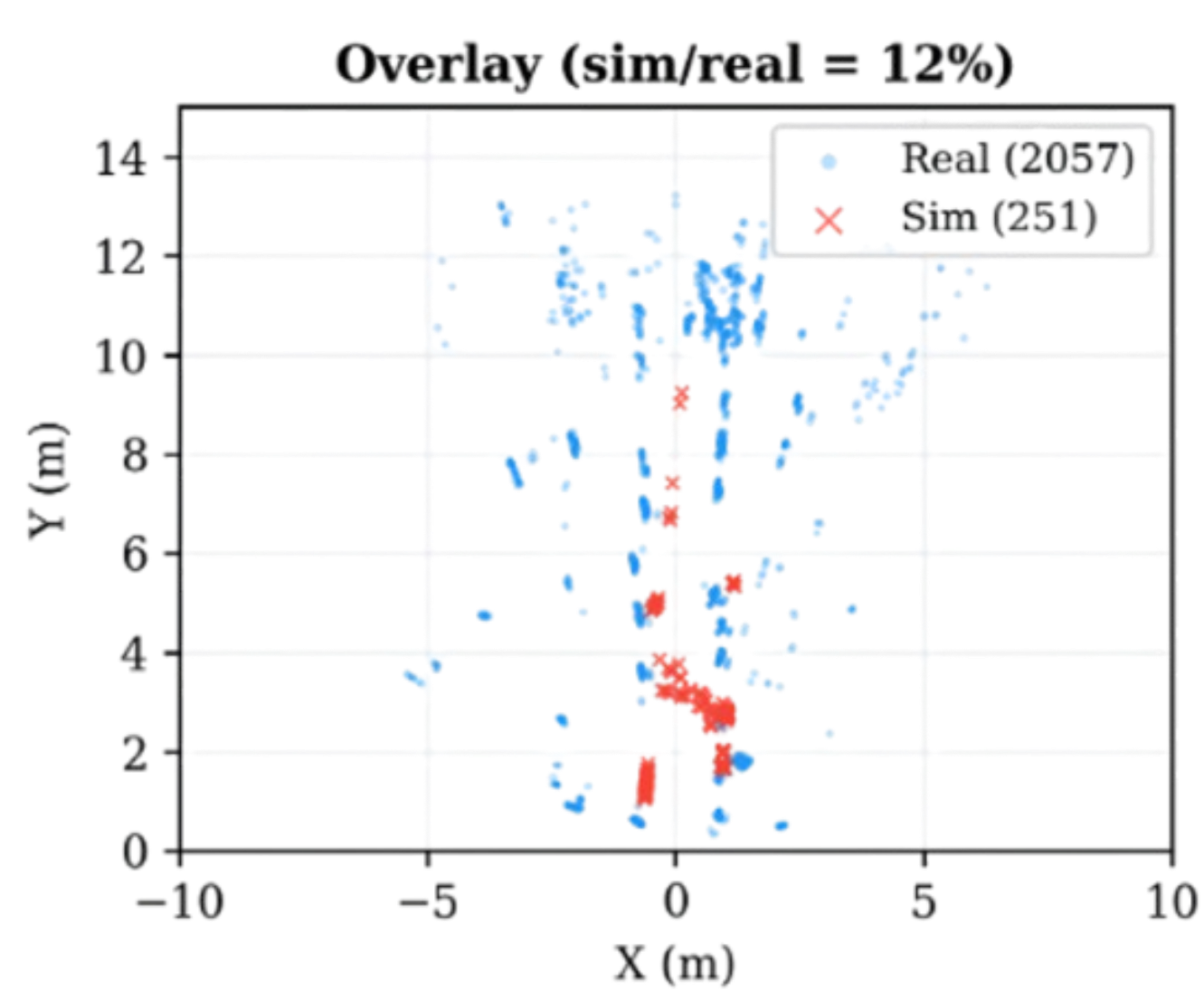}
\caption{\textbf{Sim-real radar comparison.} Blue: real radar (2,057 points). Red: simulated radar (251 points). Density ratio: 12\%. Despite density differences, both capture similar spatial structure including walls and doors.}
\label{fig:simreal}
\end{figure}

Figure~\ref{fig:simreal} compares simulated and real radar for the same scene.

\textbf{Point density gap.} Simulation produces approximately 12\% of real sensor point density (251 vs 2,057 points per frame on average). This gap arises because our ray tracer models specular reflections and first-order scattering, while real radar includes diffuse scattering, higher-order multipath, and sensor-level processing.

\textbf{Preserved structure.} Despite density differences, both simulated and real radar capture similar spatial structure. Walls appear as linear features at correct distances. Doors produce characteristic reflection patterns.

\section{TRANSFER LEARNING VIA PRE-TRAINING}

Given the substantial sim-real gap, we investigate pre-training followed by fine-tuning as a mechanism for transferring representations.

\subsection{Approach}

We use two-stage transfer: first pre-train the encoder on simulated data, then fine-tune on real data. Pre-training initializes the encoder with spatial representations learned from physically correct 3D geometry. Walls, doors, and room surfaces appear at correct positions in the simulated point clouds. Fine-tuning then adapts to real sensor characteristics while retaining this spatial knowledge. We transfer only encoder weights; the detection head is learned entirely from real data.

The value of simulation is not in data volume but in providing geometrically correct training signal. Simulated point clouds, despite being sparser than real data, place returns at physically accurate 3D positions derived from ray tracing on reconstructed meshes. This teaches the encoder where objects are in space before it encounters real sensor noise. As we show in Section 6, the benefit persists even when sim frames are fewer than real frames (Sim:Real = 0.7:1 at 100\% data), indicating that simulation provides complementary geometric information rather than additional data volume.

\subsection{Hypothesis}

Simulation, despite distributional differences, may capture structure that transfers. A radar encoder trained on simulated data learns: (1) how to aggregate spatial point patterns into meaningful features, (2) what objects look like in radar, and (3) how to extract bird's-eye-view structure from 3D points.

\section{EXPERIMENTAL EVALUATION}

We evaluate whether pre-training on simulated radar data improves 3D object detection when fine-tuning on real radar data. Our experiments test this across varying amounts of real training data to understand when simulation is most beneficial.

\subsection{Experimental Setup}

\textbf{Dataset.} We use the IFR dataset~\cite{duan2025ifr}, which provides synchronized RGB, LiDAR, and 4D mmWave radar across 10 indoor buildings with 3D bounding box annotations. The radar is a TI 76-81 GHz MMWCAS-RF cascade imaging radar (12 TX, 16 RX antennas, 0.038m range resolution, 1.18$^\circ$ azimuth resolution). We use 617 frames split into 371 train / 92 val / 155 test. Real radar produces 2,057 points per frame on average. We select IFR because it provides ground-truth 3D bounding boxes for indoor scenes, enabling quantitative evaluation of detection accuracy.

\textbf{Simulated Data.} We generate 270 simulated radar frames from 5 indoor scenes using our Mitsuba-based ray tracer configured to match IFR radar parameters. Simulated frames produce 251 points per frame (12\% of real density). We apply histogram matching in log space to align simulated intensity distributions with real radar, reducing one axis of domain shift while preserving the geometric structure we aim to transfer.

\textbf{Model.} We use PointPillars~\cite{lang2019pointpillars} implemented in OpenPCDet. The architecture consists of a PillarVFE encoder (64-dim), PointPillarScatter to produce a 128$\times$208 bird's-eye-view grid, a BaseBEVBackbone with 3-scale feature pyramid, and an AnchorHeadSingle for detection. We use a voxel size of 0.05$\times$0.05$\times$4.0m and a point cloud range of [-1.0, -5.2, -1.5] to [5.4, 5.2, 2.5]m. We select PointPillars as it is a widely-used baseline for 3D detection from point clouds.

\textbf{Training Protocol.} We compare two conditions:
\begin{itemize}
    \item \textbf{Baseline:} Train from random initialization on real data for 200 epochs.
    \item \textbf{Sim Pretrain:} Pre-train encoder on 270 sim frames for 20 epochs, then fine-tune on real data for 200 epochs.
\end{itemize}
Both use AdamW (lr=0.003), batch size 8, and full augmentation (ground-truth sampling, random flip, rotation, scaling) during fine-tuning. We use 20 pre-training epochs to learn geometric structure without overfitting to simulation-specific artifacts. We transfer only encoder weights (PillarVFE, PointPillarScatter, BaseBEVBackbone) because the encoder learns spatial representations that should transfer, while the detection head must learn class-specific predictions from real data.

\textbf{Evaluation Protocol.} We vary the amount of real training data: 5\%, 10\%, 25\%, 50\%, and 100\% of the 371 training frames. This tests whether simulation helps more when real data is scarce. For each setting, we run 5 trials with different random initializations to measure variance. We report 3D AP at IoU 0.3 on 155 test frames using the KITTI protocol. We use IoU 0.3 because radar point clouds are sparser and noisier than LiDAR, making precise localization more challenging.

\textbf{Key Comparison.} At each training data level, both baseline and sim-pretrained models see identical real training data for the same number of epochs. The only difference is encoder initialization: random vs. sim-pretrained weights. This isolates the effect of simulation pre-training.

\subsection{Results}

\begin{table}[t]
\caption{3D AP (\%) at IoU 0.3. Mean $\pm$ std across 5 runs. Sim pre-training improves 3D AP across all training data availability levels.}
\label{tab:results}
\centering
\begin{tabular}{rccccc}
\toprule
\textbf{Real Data} & \textbf{N} & \textbf{Sim:Real} & \textbf{Baseline} & \textbf{Sim Pretrain} & \textbf{$\Delta$} \\
\midrule
5\%  & 18  & 15:1  & $6.8 \pm 3.5$  & $10.1 \pm 8.4$ & $+3.3$ \\
10\% & 37  & 7.3:1 & $15.1 \pm 3.6$ & $16.0 \pm 8.2$ & $+0.9$ \\
25\% & 92  & 2.9:1 & $33.1 \pm 7.8$ & $36.3 \pm 2.9$ & $+3.2$ \\
50\% & 185 & 1.5:1 & $48.7 \pm 7.1$ & $49.9 \pm 7.0$ & $+1.2$ \\
100\% & 371 & 0.7:1 & $63.9 \pm 3.3$ & $67.5 \pm 3.4$ & $+3.7$ \\
\bottomrule
\end{tabular}
\end{table}

Table~\ref{tab:results} shows results across all training data levels. Sim pre-training improves mean 3D AP in every case, from +0.9 points at 10\% to +3.7 points at 100\%.

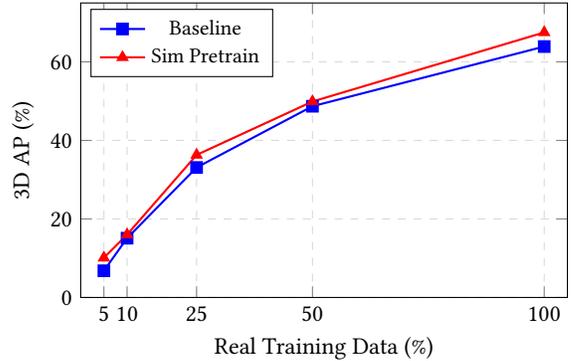
\begin{figure}[t]
\centering
\begin{tikzpicture}
\begin{axis}[
    width=0.95\columnwidth,
    height=5.5cm,
    xlabel={Real Training Data (\%)},
    ylabel={3D AP (\%)},
    xmin=0, xmax=105,
    ymin=0, ymax=75,
    xtick={5,10,25,50,100},
    legend style={
        font=\small,
        at={(0.02,0.98)},
        anchor=north west,
    },
    grid=major,
    grid style={dashed,gray!30},
]
\addplot[blue, mark=square*, thick] coordinates {(5, 6.8) (10, 15.1) (25, 33.1) (50, 48.7) (100, 63.9)};
\addplot[red, mark=triangle*, thick] coordinates {(5, 10.1) (10, 16.0) (25, 36.3) (50, 49.9) (100, 67.5)};
\legend{Baseline, Sim Pretrain}
\end{axis}
\end{tikzpicture}
\caption{Sim pre-training consistently improves 3D AP across all data regimes.}
\label{fig:results}
\end{figure}

\subsection{Analysis}

\textbf{Finding 1: Sim pre-training consistently improves 3D localization.} Across all training data levels and runs, sim pre-training yields positive mean improvement (Table~\ref{tab:results}). The gain ranges from +0.9 to +3.7 points depending on data availability.

\textbf{Finding 2: Benefits appear across all data regimes.} Gains are substantial at both low data (5\%: +3.3, 25\%: +3.2) and full data (100\%: +3.7). The benefit does not diminish as real data increases, suggesting simulation provides complementary geometric information rather than merely compensating for data scarcity.

\textbf{Finding 3: Sim pre-training reduces variance.} At 25\% data, baseline std is 7.8 vs. 2.9 for sim-pretrained. The geometric prior provides more stable initialization, reducing sensitivity to which specific frames are sampled.

\begin{figure*}[t]
    \centering
    \includegraphics[width=0.9\textwidth]{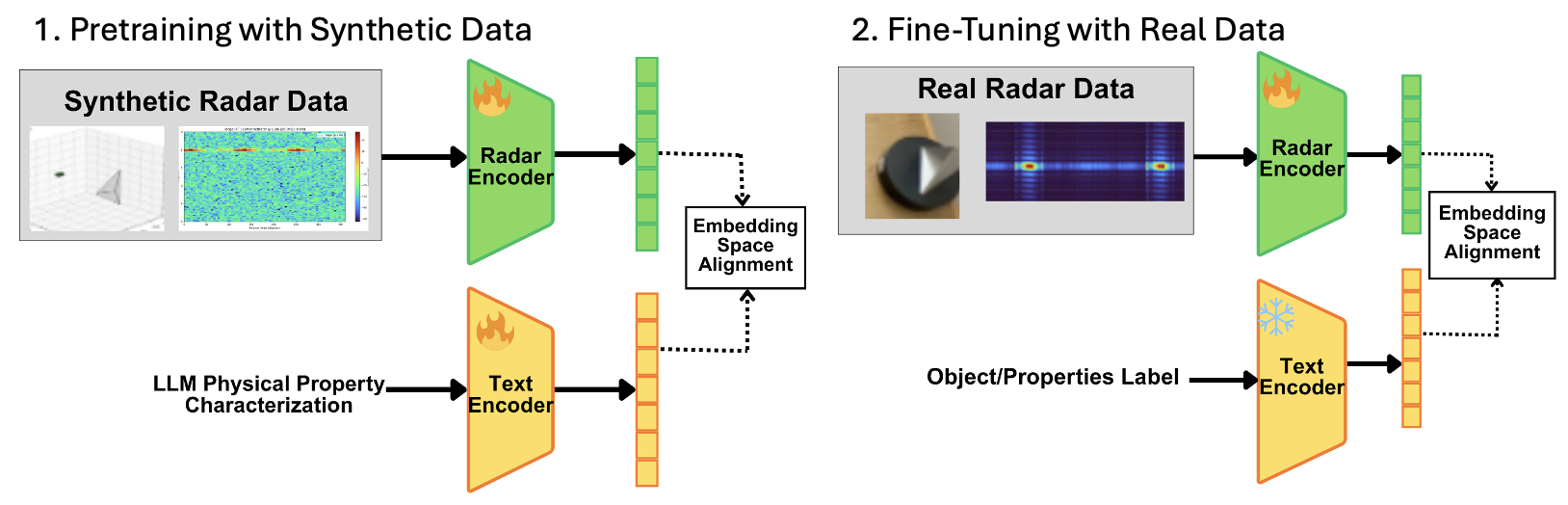}
    \caption{Proposed contrastive alignment framework for future work, where a radar encoder trained on synthetic data is aligned with text-based physical descriptions and later adapted to real radar data in a shared embedding space.}
    \label{fig:future_encoder}
\end{figure*}

\section{DISCUSSION}

\subsection{Why Does Pre-Training Help?}

The sim-real gap is substantial: simulation produces only 12\% of real point density (251 vs 2,057 points per frame). Pre-training succeeds because it initializes the encoder with useful representations. The model learns how to aggregate points within spatial cells into meaningful pillar features, how to extract spatial structure from bird's-eye-view representations, and what doors look like in radar point clouds.

\subsection{Role Of VLMs}

VLMs enable semantic material reasoning. They infer that industrial doors are metal (fire code) and floor tiles are ceramic (building conventions). This reasoning is not possible with texture-based classification.

\subsection{Limitations}

Several limitations should be noted. Our evaluation focuses on a single object class (doors) with limited simulated data (270 frames) in indoor corridor environments. The 12\% point density ratio indicates room for simulation improvement through more sophisticated scattering models. A significant limitation is field-of-view mismatch between simulation and real radar. The real radar has 120 degree azimuth coverage, but the camera-derived mesh only covers approximately 65 degrees. The effective simulation coverage is further reduced to approximately 31 degrees because geometry does not exist outside the camera view. Potential mitigations include multi-view capture, LiDAR-based mesh reconstruction, or panoramic depth estimation. While LiDAR provides metric scale for depth estimation, it is not strictly required. Monocular depth estimation methods can provide scale, though this introduces additional variability.

\subsection{Future Work}

The current evaluation focuses on indoor navigation scenarios with large static objects such as doors and walls. Future work will extend to household object detection, where targets are smaller everyday items rather than structural elements. We recorded a dataset of home object-level radar data with the simulation pipeline nearly complete for this domain. As illustrated in Fig.~\ref{fig:future_encoder}, future work will investigate a contrastive alignment framework in which radar encoders trained on synthetic data are aligned with text-based physical descriptions and subsequently adapted to real radar data in a shared embedding space. Since simulated and real radar arise from the same physical generative factors (geometry and materials) despite differing in appearance, contrastive objectives can filter out domain-specific artifacts while preserving task-relevant physical structure.

\section{CONCLUSION}

This paper presents Sim2Radar, an end-to-end framework for generating synthetic radar training data from RGB images. The system addresses a key bottleneck in radar perception: the scarcity of labeled datasets due to expensive hardware and tedious annotation requirements. Our approach combines monocular depth estimation, automatic segmentation, and vision-language material classification to reconstruct 3D scenes suitable for physics-based radar simulation. The use of VLMs for material inference enables semantic reasoning about object composition that texture-based methods cannot achieve. Experimental evaluation on the IFR dataset demonstrates that simulation pre-training improves 3D object detection performance even when the simulated data differs substantially from real sensor measurements. Pre-training on simulated frames followed by fine-tuning on limited real data yields improvements of +0.9 to +3.7 points in 3D AP, with gains concentrated in spatial localization rather than detection confidence. These results suggest that physics-based simulation provides useful priors for radar perception, particularly in low-data regimes.

\section{ACKNOWLEDGMENT}

This research was partially supported by COGNISENSE, one of seven centers in JUMP 2.0, a Semiconductor Research Corporation (SRC) program sponsored by DARPA, as well as the National Science Foundation under Grant Number CNS-1943396. The views and conclusions contained here are those of the authors and should not be interpreted as necessarily representing the official policies or endorsements, either expressed or implied, of Columbia University, NSF, SRC, DARPA, or the U.S. Government or any of its agencies.



\begin{thebibliography}{17}

\bibitem{bialer2024radsimreal}
O.~Bialer and Y.~Haitman.
\newblock {RadSimReal}: Bridging the gap between synthetic and real data in radar object detection with simulation.
\newblock In {\em CVPR}, pages 15407--15416, 2024.

\bibitem{hoydis2023sionna}
J.~Hoydis, F.~A{\"\i}t~Aoudia, S.~Cammerer, M.~Nimier-David, N.~Binder, G.~Marcus, and A.~Keller.
\newblock {Sionna RT}: Differentiable ray tracing for radio propagation modeling.
\newblock In {\em IEEE Globecom Workshops}, pages 317--321, 2023.

\bibitem{srivastava2025cshenron}
P.~Mishra, S.~Srivastava, J.~Li, K.~Bansal, and D.~Bharadia.
\newblock Demo abstract: {C-Shenron}: A realistic radar simulation framework for {CARLA}.
\newblock In {\em SenSys}, pages 726--727, 2025.

\bibitem{amatare2024dtradar}
S.~Amatare, G.~Singh, R.~Shakya, A.~Kharel, A.~Alkhateeb, and D.~Roy.
\newblock {DT-RaDaR}: Digital twin assisted robot navigation using differential ray-tracing.
\newblock {\em arXiv preprint arXiv:2411.12284}, 2024.

\bibitem{wang2025genairf}
L.~Wang, C.~Zhang, Q.~Zhao, H.~Zou, S.~Lasaulce, G.~Valenzise, Z.~He, and M.~Debbah.
\newblock Generative {AI} for {RF} sensing in {IoT} systems.
\newblock {\em IEEE Internet of Things Magazine}, 8(2):112--120, 2025.

\bibitem{chen2023rfgenesis}
X.~Chen and X.~Zhang.
\newblock {RF-Genesis}: Zero-shot generalization of mmWave sensing through simulation-based data synthesis and generative diffusion models.
\newblock In {\em SenSys}, pages 28--42, 2023.

\bibitem{chi2024rfdiffusion}
G.~Chi, Z.~Yang, C.~Wu, J.~Xu, Y.~Gao, Y.~Liu, and Y.~He.
\newblock {RF-Diffusion}: Radio signal generation via time-frequency diffusion.
\newblock In {\em MobiCom}, pages 77--92, 2024.

\bibitem{he2023fusang}
G.~He, S.~Chen, D.~Xu, X.~Chen, Y.~Xie, X.~Wang, and D.~Fang.
\newblock Fusang: Graph-inspired robust and accurate object recognition on commodity mmWave devices.
\newblock In {\em MobiSys}, pages 489--502, 2023.

\bibitem{cao2024mmclip}
Q.~Cao, H.~Xue, T.~Liu, X.~Wang, H.~Wang, X.~Zhang, and L.~Su.
\newblock {mmCLIP}: Boosting mmWave-based zero-shot {HAR} via signal-text alignment.
\newblock In {\em SenSys}, pages 184--197, 2024.

\bibitem{duan2025ifr}
K.~Duan, Z.~Zhu, and Z.~Zou.
\newblock Indoor {FireRescue} radar: {4D} indoor millimeter wave dataset and analysis for hazardous environment perception.
\newblock In {\em IROS}, pages 18620--18627, 2025.

\bibitem{wang2025moge}
R.~Wang et~al.
\newblock {MoGe}: Unlocking accurate monocular geometry estimation.
\newblock In {\em CVPR}, 2025.

\bibitem{ravi2024sam2}
N.~Ravi, V.~Gabeur, Y.-T.~Hu, R.~Hu, C.~Ryali, T.~Ma, H.~Khedr, R.~R{\"a}dle, C.~Rolber, L.~Gustafson, E.~Mintun, J.~Pan, K.~V.~Alwala, N.~Carion, C.-Y.~Wu, R.~Girshick, P.~Doll{\'a}r, and C.~Feichtenhofer.
\newblock {SAM} 2: Segment anything in images and videos.
\newblock {\em arXiv preprint arXiv:2408.00714}, 2024.

\bibitem{liu2023groundingdino}
S.~Liu, Z.~Zeng, T.~Ren, F.~Li, H.~Zhang, J.~Yang, C.~Li, J.~Yang, H.~Su, J.~Zhu, and L.~Zhang.
\newblock Grounding {DINO}: Marrying {DINO} with grounded pre-training for open-set object detection.
\newblock {\em arXiv preprint arXiv:2303.05499}, 2023.

\bibitem{chen2024internvl}
Z.~Chen, J.~Wu, W.~Wang, W.~Su, G.~Chen, S.~Xing, M.~Zhong, Q.~Zhang, X.~Zhu, L.~Lu, B.~Li, P.~Luo, T.~Lu, Y.~Qiao, and J.~Dai.
\newblock {InternVL}: Scaling up vision foundation models and aligning for generic visual-linguistic tasks.
\newblock In {\em CVPR}, 2024.

\bibitem{chow2025physbench}
W.~Chow et~al.
\newblock {PhysBench}: Benchmarking and enhancing vision-language models for physical world understanding.
\newblock {\em arXiv preprint arXiv:2501.16411}, 2025.

\bibitem{jakob2022mitsuba3}
W.~Jakob, S.~Speierer, N.~Roussel, M.~Nimier-David, D.~Vicini, T.~Zeltner, B.~Nicolet, M.~Crespo, V.~L{\'e}cuyer, and T.~Hachisuka.
\newblock Mitsuba 3: A retargetable forward and inverse renderer.
\newblock {\em ACM Transactions on Graphics}, 41(6):1--17, 2022.

\bibitem{lang2019pointpillars}
A.~H.~Lang, S.~Vora, H.~Caesar, L.~Zhou, J.~Yang, and O.~Beijbom.
\newblock {PointPillars}: Fast encoders for object detection from point clouds.
\newblock In {\em CVPR}, pages 12697--12705, 2019.

\end{thebibliography}
\end{document}